\documentclass[letterpaper, 10 pt, conference]{ieeeconf}  

\IEEEoverridecommandlockouts                              

\usepackage{xcolor}
\usepackage{hyperref}
\usepackage[normalem]{ulem}
\usepackage{graphicx}
\usepackage{amsmath,amssymb,amsfonts}
\newcommand{\bfx}{\mathbf{x}}
\newcommand{\bfz}{\mathbf{z}}
\newcommand{\bftheta}{\boldsymbol\theta}

\DeclareMathOperator*{\argmin}{arg\,min}

\newcommand{\fakest}[1]{\unskip}
\newcommand{\revision}{\textcolor{black}}
\usepackage{xcolor}
\usepackage{color,soul}

\title{\LARGE \bf
Learning to Swarm with Knowledge-Based Neural Ordinary Differential Equations}

\author{Tom Z. Jiahao$^*$, Lishuo Pan$^*$ and M. Ani Hsieh

\thanks{This work was supported by ARL DCIST CRA W911NF-17-2-0181 and Office of Naval Research (ONR) Award No. 14-19-1-2253.}
\thanks{All authors in this work are with the GRASP Laboratory, University of Pennsylvania, Philadelphia, PA 19104, USA.
        {\tt\footnotesize \{zjh, panls, m.hsieh\}@seas.upenn.edu}}%
\thanks{$^{*}$Equal contribution.}%
}

\begin{document}

\maketitle
\thispagestyle{empty}
\pagestyle{empty}

\begin{abstract}
Understanding \revision{decentralized} dynamics from collective behaviors in \fakest{natural} swarms is crucial for informing robot controller designs in artificial swarms and multiagent robotic systems. However, the complexity in agent-to-agent interactions and the decentralized nature of most swarms pose a significant challenge to the extraction of single-robot control laws from global behavior. In this work, we consider the important task of learning decentralized single-robot controllers based solely on the state observations of a swarm's trajectory. We present a general framework by adopting knowledge-based neural ordinary differential equations (KNODE) -- a hybrid machine learning method capable of combining artificial neural networks with known agent dynamics. Our approach distinguishes itself from most prior works in that we do not require action data for learning. We apply our framework to two different flocking swarms in 2D and 3D respectively, and demonstrate efficient training by leveraging the graphical structure of the swarms' information network. We further show that the learnt single-robot controllers can not only reproduce flocking behavior in the original swarm but also scale to swarms with more robots.
\end{abstract}

\section{Introduction}
Many natural swarms exhibit mesmerizing collective behaviors, and have fascinated researchers over the past decade \cite{OKUBO19861, Flierl1999-zs, Warburton1991-gq, 10.2307/1930099, Vicsek2001AQO}. A leading question is how do these global behaviors emerge from local interactions. Such fascination has led to much developments in artificial swarms and multi-agent robotic systems to emulate the swarms in nature. \cite{reynolds1987flocks, tanner2003stable, selfAssembly}. \revision{Central to these developments is the task of single-robot swarm controller synthesis, which has enabled deployment of robot swarms that respects task specifications and real-world constraints.}

Some of the earliest works on developing swarm controllers rely heavily on physical intuitions and design  controllers in a bottom-up fashion. Boids was developed by combining rules of cohesion, alignment, and separation to mimic the flocking behavior in natural swarms \cite{reynolds1987flocks}. \revision{Self-driven particles were used to model the emergence of collective behaviors in biologically motivated swarms \cite{Selfdrivenparticles}.  Flocking controllers with provably correct stability guarantees have also been developed for swarms with fixed and dynamic communication network topologies \cite{tanner2003stable, tannerpartii}.} These early works laid the foundation of decentralized swarm control and offered a glimpse of the myriad of possible swarm behaviors achievable using local single-agent controllers.

\fakest{\textit{Related Works }}In recent years, deep learning has enabled pattern discovery from complex and high-dimensional data sets. \revision{The use of neural networks (NNs) have shown promising results in a wide range of applications owing to their expressive power. This has opened up potential avenues for data-driven learning of single-robot swarming control strategies in more efficient and scalable ways. In this work, we leverage recent advances in scientific machine learning and employ  knowledge-based neural ordinary differential equations (KNODE) \cite{Jiahao2021Knowledgebased} for learning swarm controllers directly from observations of a swarm. We demonstrate that through our top-down approach to controller synthesis, global behaviors of different swarms can be successfully reproduced based on the past observations of their evolution.}

\section{\revision{Related Works}}
Various data-driven methods have been used to model local control policy in swarms. Feedforward neural networks have been used to approximate decentralized control policies by training on the observation-action data from a global planner \cite{Riviere_2020}. Furthermore, deep neural networks have been used to model higher order residual dynamics to achieve stable control in a swarm of quadrotors \cite{shi2020neural}. Recently, graph neural networks (GNN) have been extensively used in swarms, owing to their naturally distributed architecture. GNN allows efficient information propagration through networks with underlying graphical structures \cite{tolstaya2020learning}, and have been noted for their stability and permutation equivariance \cite{gama2021graph}. Decentralized GNN controllers have been trained with global control policies to imitate \revision{swarm behaviors \cite{tolstaya2020learning, 2019clone}}. All these works pose the controller synthesis problem as an imitation learning problem, and require knowledge of the actions resulting from an \textit{optimal} control policy for learning or improving the local controllers. In practice, action data can be difficult to access, especially when learning behaviors from natural or adversarial swarms. In addition, GNNs can potentially allow a robot to access the state information of robots outside its communication range through information propagation. The \revision{true} extent of decentralization may therefore be limited when more propagation hops are allowed.

Deep reinforcement learning has also been applied to swarms for various applications \cite{huttenrauch2019deep}. Early works like \cite{huttenrauch2017guided} learn a decentralized control policy for maintaining distances within a swarm and target tracking.
An inverse reinforcement learning algorithm was presented in \cite{vsovsic2016inverse} to train a decentralized policy by updating the reward function alongside the control policy based on an expert \fakest{policy} \revision{behavior}. In addition, GNNs have also been used within the reinforcement learning framework for learning connectivity for data distribution \cite{tolstaya2021learningconnectivity}. However, reinforcement learning is usually employed to solve task-specific problems with well-defined goals \revision{and need to tackle the challenge of reward shaping}. The specific objectives of swarms may be difficult to discern from only observations, and therefore reinforcement learning is often not suitable for \fakest{learning control strategies} \revision{learning global behaviors} from solely observational data.

The contribution of this work is three-fold. First, we demonstrate the feasibility of learning single-robot controllers that can achieve the observed global swarming behaviors from only swarm trajectory data. Second, we propose a generalized model for incorporating known robot dynamics to facilitate learning single-robot controllers. Lastly, we show how to efficiently scale KNODE for learning from local information in a multi-agent setting.

\section{Problem Formulation}
We consider the problem of learning single-robot controllers based on the observations of the trajectory of a swarm. We assume that the swarm is homogeneous, {\it i.e.}, all robots in the swarm use the same controller. Given a swarm of $n$ agents, we make $m$ observations at sampling times $T = \{t_1, t_2,...,t_m\}, t_i\in \mathbb{R}$ given by
\begin{equation*}
\left[\begin{array}{c}
\mathbf{Z}^T\left({t_1}\right) \\
\mathbf{Z}^T\left({t_2}\right) \\
\vdots \\
\mathbf{Z}^T\left({t_m}\right)
\end{array}\right]=\left[\begin{array}{cccc}
\bfz_{1}\left({t_1}\right) & \bfz_{2}\left({t_1}\right) & \cdots & \bfz_{n}\left({t_1}\right) \\
\bfz_{1}\left({t_2}\right) & \bfz_{2}\left({t_2}\right) & \cdots & \bfz_{n}\left({t_2}\right) \\
\vdots & \vdots & \ddots & \vdots \\
\bfz_{1}\left({t_m}\right) & \bfz_{2}\left({t_m}\right) & \cdots & \bfz_{n}\left({t_m}\right)
\end{array}\right],
\end{equation*}
where the matrix $\mathbf{Z}(t_i)\in \mathbb{R}^{n\times d}$ is the observations of the states of all $n$ agents at $t_i$, and the vector $\bfz_i(t_j)\in \mathbb{R}^{d}$ is the state of agent $i$ observed at $t_j$ with dimension $d$. For instance, in a first-order system, an agent modeled as a rigid body in a 3-dimensional space has $d=6$, where the first three dimensions correspond to the positions and the last three the orientations. Our goal is to learn a single-robot controller solely from the observations $\mathbf{Z}$. \revision{Notice that control inputs are not assumed to be part of the observations.}

The evolution of each individual robot's state can be described by the true dynamics given by
\begin{equation}
\label{eqn: agent dynamics}
    \dot{\bfz}_i(t) = f_i(\bfz_i, u_i),
\end{equation}
where $\bfz_i$ is the state of robot $i$, and $u_i$ is its control law. The function $f_i(\cdot, \cdot)$ defines the dynamics given the state of robot and control law $u_i$. It is assumed that all robots in the swarm have the same dynamics and control strategy, and therefore we can drop the subscripts and rewrite \eqref{eqn: agent dynamics} as $\dot{\bfz}_i(t) = f(\bfz_i, u)$ for all $i$. The control law $u$ is a function of the states of other robots in the swarm, and defines the interaction between robot $i$. For example, \revision{a communication radius may be enforced by the control law $u$ to let each robot only interact with its neighbors. } \fakest{a control law $u$ can be designed to let each robot only interact with its neighbors within some communication radius. }

The dynamics of the entire swarm can be written as a collection of the single-robot dynamics as
\begin{equation}
    \dot{\mathbf{Z}}(t) = [\dot{\bfz}_1(t), \dot{\bfz}_2(t), \cdots, \dot{\bfz}_n(t)]^T.
\end{equation}

Given the initial conditions of all robots $\mathbf{Z}_0$ at $t_0$, the states of all robots at $t_1$ is given by
\begin{equation}
    \label{eqn: swarm int}
    \mathbf{Z}(t_1) = \mathbf{Z}_0 + \int^{t_1}_{t_0} \dot{\mathbf{Z}}(t) dt.
\end{equation}

In practice, the integration in \eqref{eqn: swarm int} is performed numerically. Our task is to find a single-robot control law parameterized by $\bftheta$ as part of the single-robot dynamics given by
\begin{equation}
    \label{eqn: u hat}
    \dot{\bfz}_i(t) = \hat{f}(\bfz_i, \hat{u}_{\bftheta}),
\end{equation}
where $\hat{u}_{\bftheta}$ is the single-robot control law parameterized by $\bftheta$. The learnt controller should best reproduce the observed \revision{global} swarm behaviors. \revision{Note that the high dimensionality of a swarming system means that similar collective dynamics can be achieved with very disparate collections of single-robot trajectories.  This suggests that it may be impractical to predict each individual trajectory in a swarm over long time horizons. Instead, we focus on learning and reproducing the global behaviors of swarms based on metrics, which we will formalize in later sections.}

\section{Knowledge-based Neural Ordinary Differential Equations (KNODE)}

KNODE is a scientific machine learning framework that applies to a general class of dynamical systems. It has been shown to model a wide variety of systems with nonlinear and chaotic dynamics, \revision{with robustness to noise and irregularly sampled data} \revision{\cite{Jiahao2021Knowledgebased}}. In our problem, we assume a single-robot dynamics in the form of \eqref{eqn: u hat}. From a dynamical systems perspective, \fakest{$f(\bfz_i, \hat{u}_{\bftheta})$} \revision{$\hat{f}(\bfz_i, \hat{u}_{\bftheta})$} is a vector field. This makes KNODE a suitable method \fakest{for our problem} \revision{to learn $\hat{f}(\bfz_i, \hat{u}_{\bftheta})$} because it directly models vector fields using neural networks \cite{Jiahao2021Knowledgebased}. To put KNODE in the context of our learning problem, given some known swarm dynamics $\tilde{f}(\mathbf{Z})$ as knowledge, KNODE optimizes for the control law as part of a dynamics given by
\begin{equation}
    \label{eqn: knowledge incorporation}
    \dot{\bfz}_i(t) = \hat{f} (\bfz_i,  \hat{u}_{\bftheta}, \tilde{f}(\mathbf{Z})),
\end{equation}
where the control law $\hat{u}_{\bftheta}$ is a neural network, and $\hat{f}$ defines the coupling between the knowledge and the rest of the dynamics. While the original KNODE linearly couples a neural network with $\tilde{f}$ using a trainable matrix \fakest{$\mathbf{M}_{out}$} \cite{Jiahao2021Knowledgebased}, we note that the way knowledge gets incorporated is flexible. In later sections we will demonstrate how to effectively incorporate knowledge for learning single-robot controllers. Furthermore, the ability to incorporate knowledge will require less training data \cite{physicslearning, Jiahao2021Knowledgebased}.

We minimize the mean squared error (MSE) between the observed trajectories and the trajectories predicted from the estimate dynamics using $\hat{u}_{\bftheta}$ for robot $i$.  A loss function is given by
\fakest{$L(\bftheta) = \frac{1}{m-1}\sum^{m-1}_{j=1}\sum^{n}_{i=1}\|\hat{\bfz}_i(t_{j+1}, \bfz_i(t_j)) - \bfz_i(t_{j+1})\|^2,$}
\revision{
\begin{equation}
    \label{eqn: objective}
    L(\bftheta) = \frac{1}{m-1}\sum^{m-1}_{j=1}\sum^{n}_{i=1}\|\hat{\bfz}_i(t_{j+1}, \bfz_i(t_j)) - \bfz_i(t_{j+1})\|_2^2,
\end{equation}
}
where $\hat{\bfz}_i(t_{j+1}, \bfz_i(t_j))$ is the estimated state of robot $i$ at $t_{j+1}$ generated using the initial condition $\bfz_i(t_j)$ at $t_j$, and it's given by
\begin{equation}
    \label{eqn: agent int}
    \hat{\bfz}_i(t_{j+1}, \bfz_i(t_j)) = \bfz_i(t_j) + \int^{t_{j+1}}_{t_j} \hat{f} (\bfz_i,  \hat{u}_{\bftheta}, \tilde{f}(\mathbf{Z})) dt.
\end{equation}
Intuitively, the loss function in \eqref{eqn: objective} computes the one-step-ahead estimated state of all robots from every snapshot in the observed trajectory, and then computes the average MSE between the estimated and observed states for the entire trajectory \fakest{form} \revision{from} $t_{1}$ to $t_{m-1}$.

Our learning task can then be formulated as an optimization problem given by
\begin{align}
\label{eqn: optimization}
    \min_{\bftheta} \quad & L(\bftheta), \\
    \textrm{s.t.} \quad & \dot{\bfz}_i = \hat{f} (\bfz_i,  \hat{u}_{\bftheta}, \tilde{f}(\mathbf{Z})),\ \text{for all}\ i,
\end{align}
which includes the dynamics constraint for all robots in the swarm. The parameters $\bftheta$ can then be estimated by $\bftheta = \argmin_{\bftheta} L(\bftheta).$ The gradients of $\bftheta$ with respect to the loss can be computed by either the conventional backpropagation or the adjoint senesitivity method. The adjoint sensitivity method has been noted as a more memory efficient approach than backpropagation, though at the cost of training speed~\cite{hasani2020liquid}. In this work, we use the adjoint method for training similar to that in~\cite{conf/nips/ChenRBD18} and \cite{Jiahao2021Knowledgebased}.

\section{Method}
In this section, we walk through the process \fakest{to construct} \revision{for constructing} \fakest{$\hat{f}(\bfz_i, \hat{u}_{\bftheta})$} \revision{$\hat{f} (\bfz_i,  \hat{u}_{\bftheta}, \tilde{f}(\mathbf{Z}))$} in the context of learning to swarm and the incorporation of knowledge in the form of known single robot dynamics.

\subsection{Decentralized Information Network} 
We assume a robot in a swarm can only use its local information as inputs to its controller. To incorporate this assumption, we impose a decentralized information network on the swarm. Specifically, we assume robots have finite communication \revision{radii as denoted by $d_{cr}$.}\fakest{ranges and} In addition, each robot can only communicate with a maximum number of neighbors\revision{, including itself, as} denoted by $k$. \fakest{within this range. We denote the communication radius by $d_{cr}$.} We refer to \fakest{these neighbors} \revision{the robots within this radius} as the \textit{active neighbors}. If there are more than $k$ neighbors within a robot's communication radius, \fakest{only} the closest $k$ neighbors are considered to be \fakest{the} {\it active} \fakest{ones}.

We leverage the communication graph of the swarm to compute the local information for each robot at each time step. The communication graph at time $t$ can be described by a \textit{graph shift operator} $\mathbf{S}(t) \in \mathbb{R}^{n\times n}$, which is a binary adjacency matrix computed based on $d_{cr}$ and the positions of all robots at each time step. In this work, we treat the communication radius $d_{cr}$ as a hyperparameter. Note that the communication graph is time-varying because the information network changes as robots move around in a swarm. Then $\mathbf{S}_{ij}(t)=1$ if the Euclidean distance between agents $i$ and $j$ is less than or equal to $d_{cr}$, and $\mathbf{S}_{ij}(t)=0$ otherwise. The index set of the neighbors of robot $i$ at time $t$ is therefore given by
\begin{equation}
    \label{eqn: neighbors}
    \mathcal{N}_i(t) = \{j |j\in \mathcal{I}, \mathbf{S}_{ij}(t)=1\},
\end{equation}
where $\mathcal{I}=1, \ldots, n$ is the index set of all robots. Note that set of neighbors of robot $i$ also includes itself. At time $t$, the information kept by robot $i$ is the matrix $\mathbf{Y}_i(t)\in \mathbb{R}^{k\times d}$ given by
\begin{equation}
    \label{eqn: dec info struct}
    \mathbf{Y}_i(t) = g(\{\bfz_j(t) | j\in \mathcal{N}_i(t)\}, k),
\end{equation}
where the function $g(\cdot, k)$ maintains the dimension of the matrix $\mathbf{Y}_i(t)$, and forms the rows of matrix $\mathbf{Y}_i(t)$ using the state information of robot $i$'s active neighbors in \fakest{descending} \revision{ascending} order of their Euclidean distance from robot $i$. \revision{Naturally, robot $i$'s state is always in the first row because its distance to itself is $0$}. If there are fewer than $k$ active neighbors within a robot's communication radius, the remaining rows in $\mathbf{Y}_i(t)$ are padded with zeros. In this work, $k$ is treated as a hyperparameter.

The matrix $\mathbf{Y}(t)$ represents the local information accessible to each robot at time $t$ and it completes the decentralized information network of the swarm. In summary, \eqref{eqn: neighbors} and \eqref{eqn: dec info struct} enforces the assumptions of finite communication and perception radii for each robot.

\subsection{Information Time Delay}
\begin{figure}
    \centering
    \vspace*{0.2cm}\includegraphics[width=0.45\textwidth]{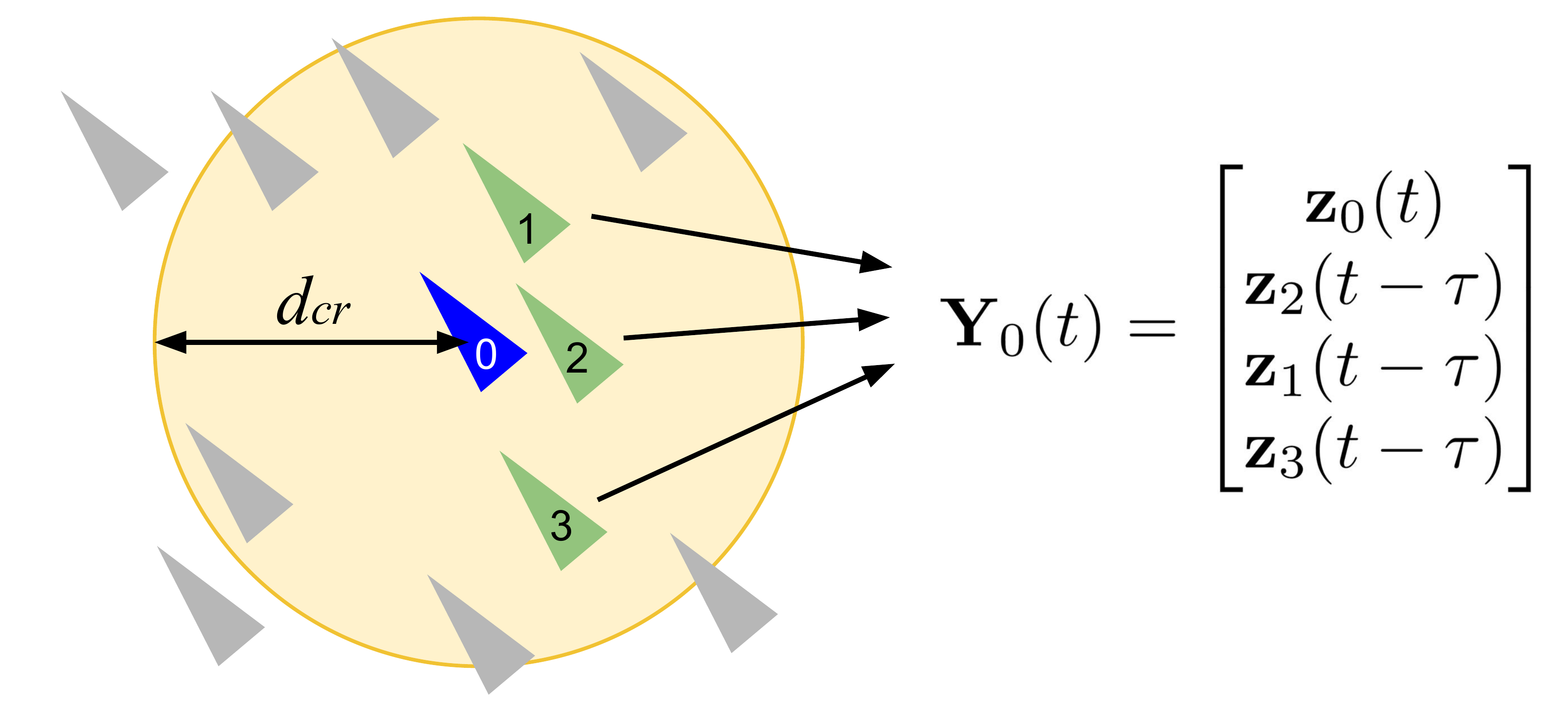}
    \caption{Decentralized information network for robot $0$ with time delay $\tau$, and $3$ active neighbors. The image shows robot $0$'s egocentric view, where 8 neighbors are within its communication range $d_{cr}$. Only the closest three neighbors contribute to the information structure of robot $0$. Their states from $t-\tau$ are ordered based on their proximity to robot $0$ to form $\mathbf{Y}_0(t)$.}
    \label{fig: dec info}
\end{figure}

In addition to a decentralized information structure, we further assume that each robot only gets delayed state information from its neighboring robots by a time lag $\tau$. This is to emulate the latency in agent communication in real swarms. With time delay, the information \revision{accessible to} \fakest{structure of} robot $i$ in \eqref{eqn: dec info struct} becomes \fakest{$\mathbf{Y}_i(t) = g(\{\bfz_j(t-\tau) | i \neq j, j\in \mathcal{N}_i(t-\tau)\}, k).$}
\revision{\begin{equation}
    \label{eqn: delay info struct}
    \mathbf{Y}_i(t) = 
    \begin{bmatrix}
    \bfz_i^T(t)\\ 
    g\left(\left\{\bfz_j(t-\tau) | i \neq j, j\in \mathcal{N}_i(t-\tau)\right\}, k-1\right)
    \end{bmatrix}.
\end{equation}
}
Fig. \ref{fig: dec info} shows an example of the information structure described by \eqref{eqn: delay info struct} using $k=3$. The process of constructing $\mathbf{Y}_i(t)$ for all $t\in T$ in \eqref{eqn: neighbors}, \eqref{eqn: dec info struct} and \eqref{eqn: delay info struct} leverages the graphical structure of the swarm's information network. During training, the collection of delayed neighbor information is done efficiently through the matrix multiplication $\mathbf{S}(t-\tau)\mathbf{Z}(t-\tau)$, which leaves for each robot only the state information of its neighbors at $t-\tau$. Then for robot $i$ we append the $i$th row of $[\mathbf{S}(t-\tau)\mathbf{Z}(t-\tau)]$ to its own state \fakest{$\bfz_i(t)$} \revision{$\bfz^T_i(t)$}. Finally we only keep $k$ rows of the resulting matrix to form $\mathbf{Y}_i(t)$. Compared to some GNN approaches \cite{tolstaya2020learning, gama2021graph}, the information structure $\mathbf{Y}_i(t)$ in our work is more explicit. A robot with GNN controllers can only access the diffused state information from other robots, $i.e.$ the neighbors' information has been repeatedly multiplied by the graph operators before reaching this robot. In this work, we directly let each robot access the state information of its active neighbors. In real-world implementation of robot swarms, our proposed information structure in \eqref{eqn: delay info struct} is more realistic as each robot can easily subscribe to or observe its neighbors' states. In addition, the information structure $\mathbf{Y}_i(t)$ enables scalable learning as we can treat the robots in a swarm as batches. As a result, training memory scales linearly with the number of robots in the swarm, and training speed scales sub-linearly.

\subsection{Knowledge Embedding}
In this work, a potential-function-based obstacle avoidance strategy similar to~\cite{potentialfunction} is used as knowledge. Let the distance between robot $i$ and an obstacle $\mathcal{O}$ be $d_\mathcal{O}(\bfz_i)$, where $\bfz_i$ is the state and includes the position of robot $i$. The potential function is then given by
\begin{equation}
\label{eqn: potential function}
  U_\mathcal{O}(\bfz_i) =
    \begin{cases}
      \frac{\lambda}{2}\frac{1}{d_\mathcal{O}^2(\bfz_i)} & \text{if } d_\mathcal{O}(\bfz_i) \leq d_0,\\
      0 & \text{otherwise},
    \end{cases}
\end{equation}
where $\lambda$ is the gain, and $d_0$ is the obstacle influence threshold (\textit{i.e.} the distance within which the potential function becomes active). Based on this potential function, the repulsive force to avoid the obstacle $\mathcal{O}$ is given by
\begin{equation}
\label{eqn: potential force}
  F_\mathcal{O}(\bfz_i) =
    \begin{cases}
      -\nabla U_\mathcal{O}(\bfz_i) & \text{if } d_\mathcal{O}(\bfz_i) \leq d_0,\\
      0 & \text{otherwise},
    \end{cases}
\end{equation}

When multiple obstacles are present, the repulsive forces computed from each obstacle are summed for a resultant repulsive force. For collision avoidance, we assume that each agent will only actively avoid its closest neighbor within $d_0$ at any given time.

Assuming that the robots in a swarm follow first-order dynamics, we combine the decentralized information network in \eqref{eqn: dec info struct} and the knowledge in \eqref{eqn: potential force} into \fakest{a} \revision{the} dynamics given by
\begin{equation}
\label{eqn: final controller}
    \dot{\bfz}_i = \hat{f}(\bfz_i,  \hat{u}_{\bftheta}(\mathbf{Y}_i, \bfz_i)) - \sum_j \lambda_j \nabla U_{\mathcal{O}_j}(\bfz_i),
\end{equation}
where $u_{\bftheta}$ is a neural network, and $\lambda_j$ is a trainable gain for avoiding obstacle $\mathcal{O}_j$.  \revision{Note that Eqn. \eqref{eqn: final controller} further illustrates how our framework differs from imitation learning. While $\dot{\bfz}_i$ and $\hat{u}_{\bftheta}$ are the learnt dynamics and control policy which drive the system, they do not have to be part of the training data.} \fakest{Both $u_{\bftheta}$ and $\lambda_j$ are trained using the KNODE framework. Note that since all robots in a homogeneous swarm share the same control law, the training is done in batches of robots to accelerate training.}

\section{Learning to flock in 2D}
We \revision{first} use a global controller proposed by \cite{tanner2003stable} to generate observations for our learning problem. 
\subsection{Simulation in 2D and training}
This global controller achieves stable flocking, which ensures eventual velocity alignment, collision avoidance and group cohesion in a swarm of robots. The robots follow the double integrator dynamics given by
\begin{equation}
\label{eqn: 2nd order dynamics}
    \begin{aligned}
    \dot{\mathbf{r}}_i &= \mathbf{v}_i,\\
    \dot{\mathbf{v}}_i &= \mathbf{u}_i,\ \ i=1,...,n,
    \end{aligned}
\end{equation}
where $\mathbf{r}_i$ is the 2D position vector of robot $i$, $\mathbf{v}_i$ is its velocity vector. The full state of each robot is therefore $\bfx = [\mathbf{r}, \mathbf{v}]\in \mathbb{R}^4$. The control law $\mathbf{u}_i$ is given by
\begin{equation}
\label{eqn: 2D global controller}
    \mathbf{u}_i = -\sum_{j\in \mathcal{N}_i} (\mathbf{v}_i - \mathbf{v}_j) - \sum_{j\in \mathcal{N}_i} \nabla_{\mathbf{r}_i} V_{ij},
\end{equation}
where $V_{ij}$ is a differentiable, nonnegative, and radially unbounded function of the distance between robot $i$ and $j$ \cite{tanner2003stable}. The first summation term in \eqref{eqn: 2D global controller} aims to align the velocity vector of robot $i$ with those of its flockmates, while the second summation term is the total potential field around robot $i$ responsible for both collision avoidance and cohesion~\cite{tanner2003stable}. The set $\mathcal{N}_i$ is the set of all robots in the swarm for the global controller. 

We use the \revision{explicit fourth-order} Runge-Kutta method to simulate the dynamics in \eqref{eqn: 2D global controller} with a \revision{step time} of 0.01. \revision{Given $n$ robots, their} \fakest{The} locations \fakest{of robots} are initialized uniformly on a disk with radius $\sqrt{n}$ to normalize the density within the swarm. The velocities of robots are initialized uniformly with magnitudes between $[0, 3]$. Additionally, a uniformly sampled velocity bias with magnitude between [0, 3] is added to the swarm. A total of \fakest{35} \revision{50} trajectories are simulated, each with a total of 2000 steps. The lengths of the trajectories is chosen such that the swarms will converge to stable flocking. We use 30 trajectories as the training data, and the remaining \fakest{5} \revision{20} as the testing data. We added zero-mean Gaussian noise with variance 0.001 to the training trajectories. This is known as \textit{stabilization noise} in modeling dynamical systems and has been shown to improve model convergence \cite{OttRescomp}.

The training model follows \eqref{eqn: final controller}. There are no obstacles to avoid in the 2D case, so the potential function is only used to avoid collision among the agents. Specifically, we let each robot avoid its closest neighbor at every time step. For the controller $\hat{u}_{\bftheta}$, we use a one layer neural network with 128 hidden units, and a hyperbolic tangent activation function. The trainable gain for collision avoidance is defined as $\lambda = a + \phi^2$, where $a$ is a positive number for setting the minimum amount of force to avoid collision. The single parameter $\phi$ is trained together with the neural network\fakest{to provide more forces for collision avoidance as needed}. We do not assume information delay in the 2D case.

\subsection{Evaluating flocking in 2D}
We evaluate 2D flocking behavior using two metrics:

\noindent \textbf{Average velocity difference} ($avd$) measures how well the velocities of robots are aligned\fakest{in 2D flocking}. It is given by
\begin{equation}
    avd(t) = \frac{2}{n(n-1)} \sum_{i \neq j} ||\mathbf{v}_{i}(t) - \mathbf{v}_{j}(t)||_{2}.
    \label{eqn: avd}
\end{equation}

\noindent \textbf{Average minimum distance to a neighbor} ($amd$) measures the cohesion between agents in both 2D and 3D when flocking is achieved. It is given by
\begin{equation}
\label{eqn: amd}
    amd(t) = \frac{1}{n} \sum_{i=1}^{n} \min\limits_{j} ||\mathbf{r}_{i}(t) - \mathbf{r}_{j}(t)||_{2}.
\end{equation}
$amd$ should decrease as the robots move closer together, but it should not reach zero if collision avoidance is in place. \revision{To generate trajectories using the learnt controller, we use it to replace \eqref{eqn: 2D global controller} in the dynamics described by Eqn. \eqref{eqn: 2nd order dynamics} for acceleration control.}

\subsection{2D Results}
\begin{figure}
    \centering
    \vspace*{0.2cm}\includegraphics[width=0.35\textwidth]{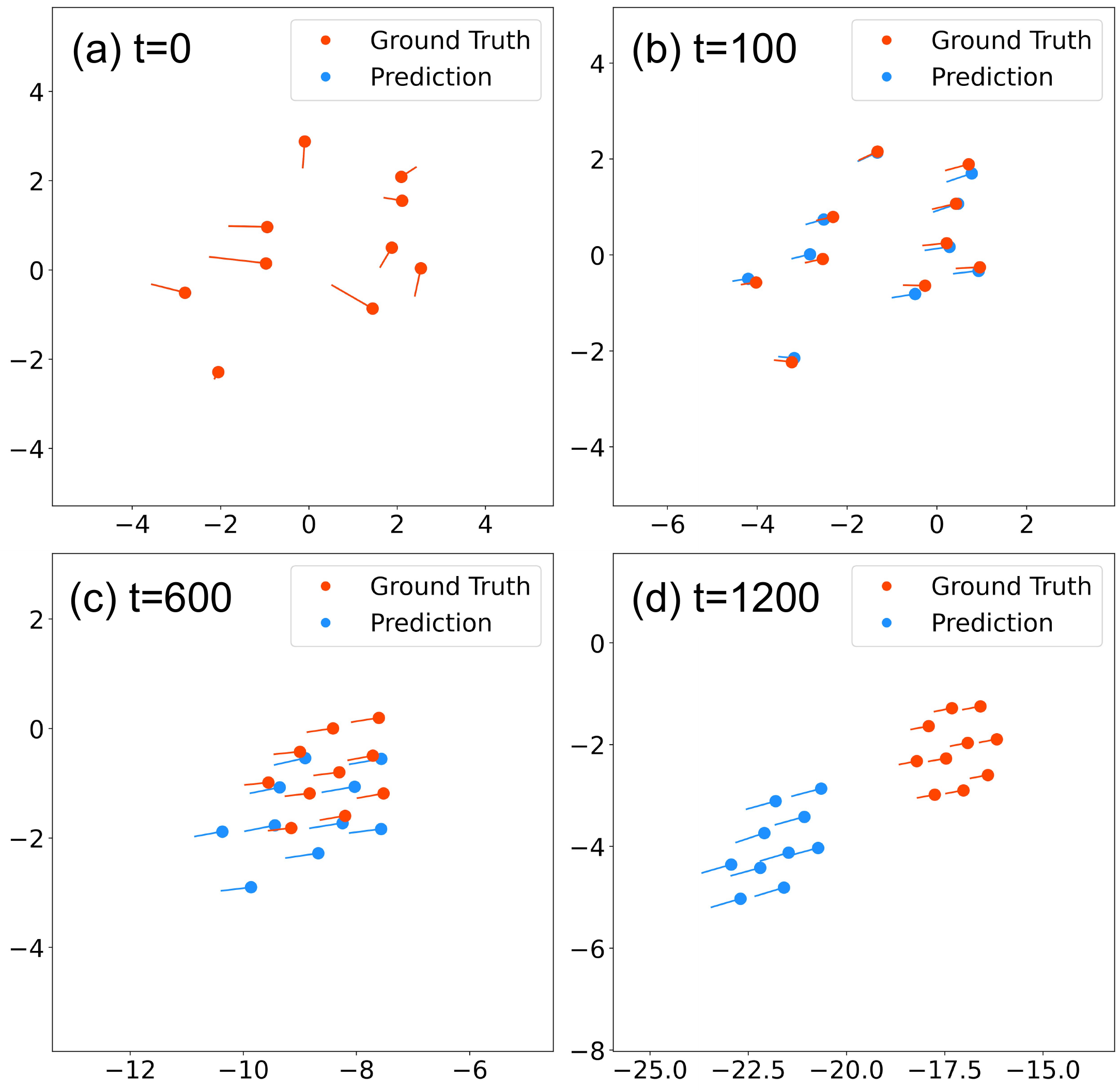}
    \caption{\revision{Predicted trajectory of 10 robots using the learnt controller ($d_{cr} = 5, k = 6$) with the same initial states as the testing trajectory (ground truth)}. The subfigures (a)(b)(c)(d) show the snapshots of the swarm at $t=0, 100, 600$, and $1200$ respectively.}
    \label{fig: 2D prediction}
\end{figure}

\begin{figure}
    \centering
    \vspace*{0.2cm}{\includegraphics[width=0.42\textwidth]{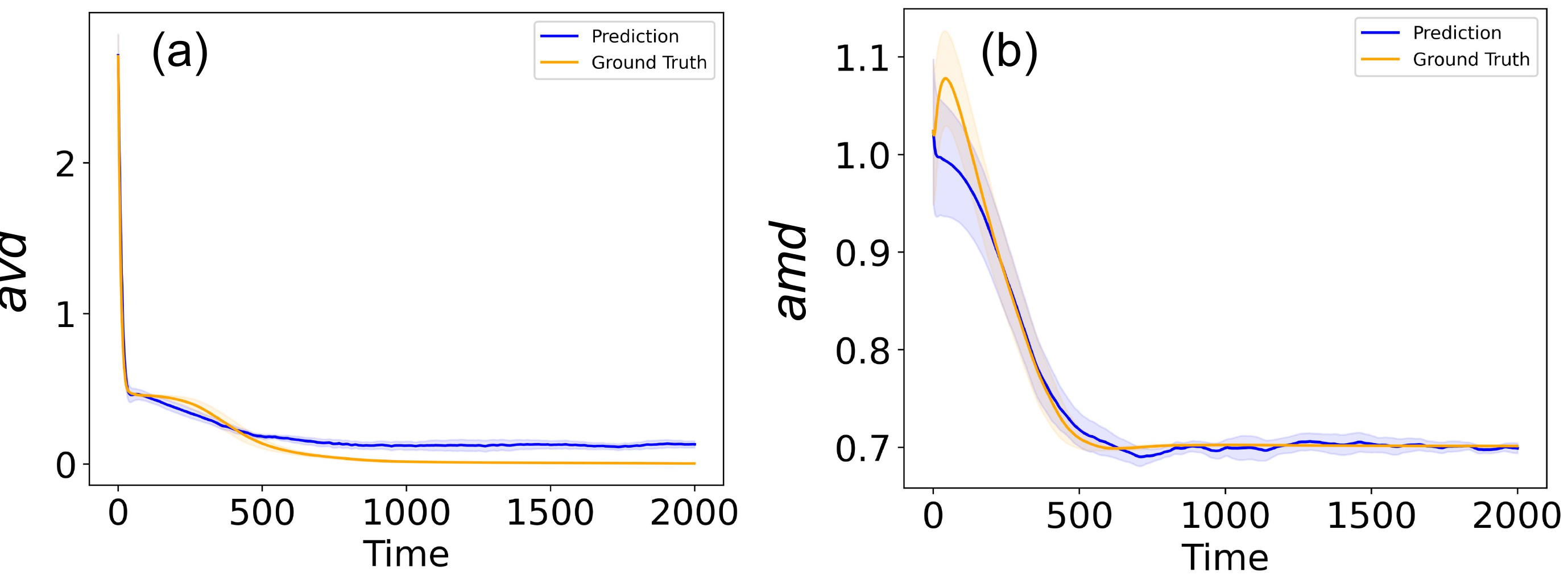}}
    \caption{The metrics for the learnt 2D controller ($d_{cr} = 5, k = 6$) show (a) average velocity difference, and (b) average minimum distance to a neighbor. The $95\%$ confidence intervals are based on 20 sets of testing trajectories.}
    \label{fig: 2D metric}
\end{figure}

Fig. \ref{fig: 2D prediction} shows four snapshots of \fakest{one} \revision{the} swarm trajectory generated using the trained single-robot controller, \revision{and provides a qualitative comparison between the prediction and ground truth}. \revision{The controller used to produce these snapshots were trained with $d_{cr} = 5$ and $k=6$}. \fakest{The communication radius is 5 and the number of active neighbors is 6.} The robots \fakest{in the snapshots} are initialized using the initial states from \fakest{one of} the testing \revision{trajectory} \fakest{trajectories}. It can be observed that \fakest{at around $t=600$, the predicted robots have mostly aligned their velocities in the same direction,} \revision{the predicted swarm achieves velocity alignment while the robots stay apart from each other,} indicating the emergence of flocking behavior. This can be further verified by the metrics for 2D flocking as shown in Fig. \ref{fig: 2D metric}. The predicted swarm \revision{trajectory} follows similar trends as the ground truth under both metrics. 

\begin{figure}
    \centering
    {\vspace*{0.2cm}\includegraphics[width=0.42\textwidth]{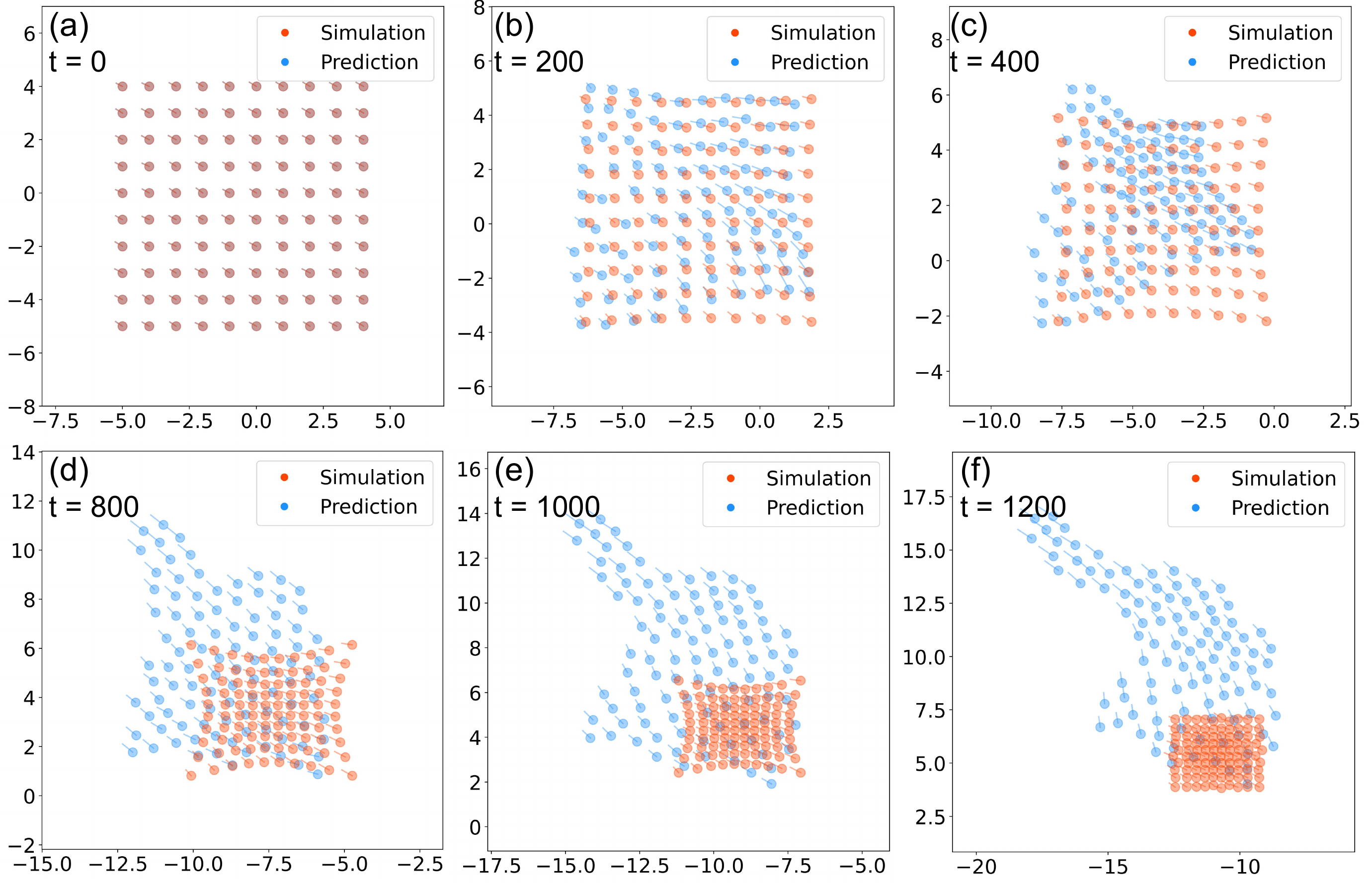}}
    \caption{Predicted trajectory of 100 robots using the learnt controller ($d_{cr} = 5, k = 6$) with uniformly initialized positions. The subfigures (a)(b)(c)(d)(e)(f) show the snapshots of the swarm at $t=0, 200, 400, 800, 1000$ and $1200$ respectively.}
    \label{fig: 2D scaling}
\end{figure}

\begin{figure}
    \centering
    \vspace*{0.1cm}{\includegraphics[width=0.45\textwidth]{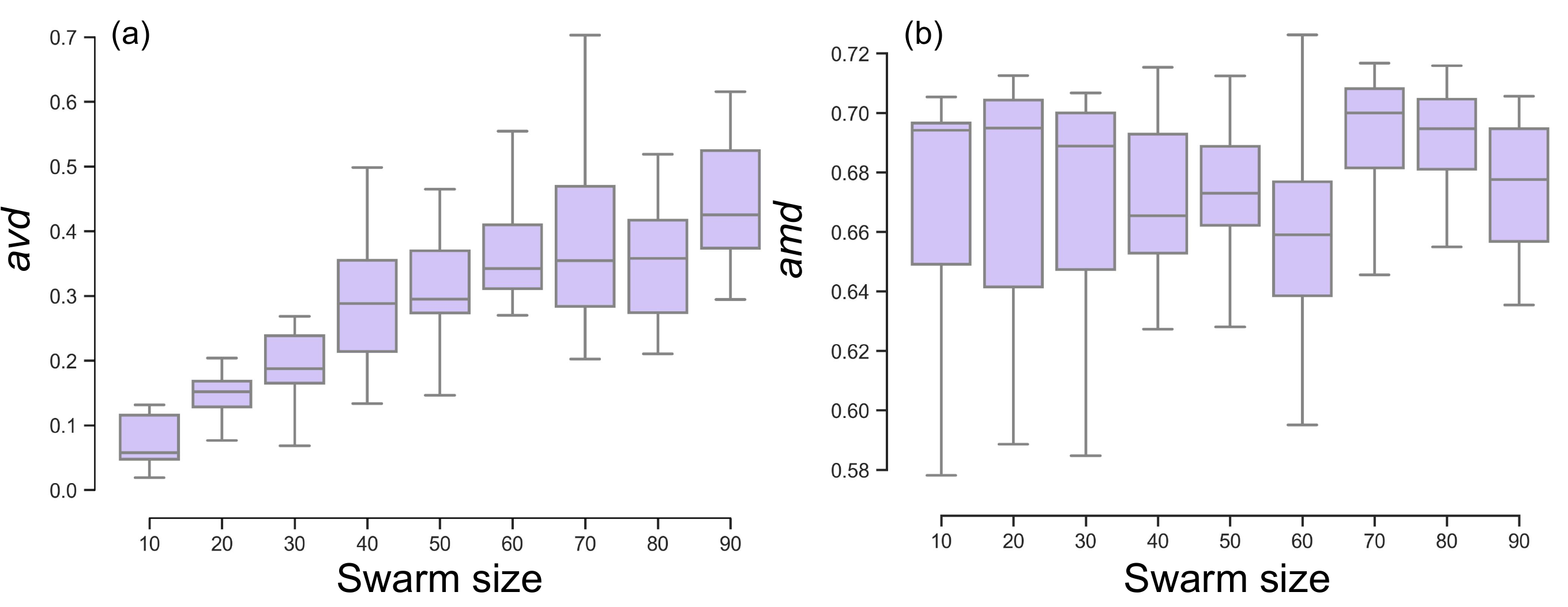}}
    \caption{\revision{Box plot of (a) average velocity difference ($avd$), and (b) average minimum distance to a neighbor ($amd$) on scaling to different swarm sizes using a trained controller in 2D. For each swarm size, the box represents the statistics of 15 runs using different initial conditions.}}
    \label{fig: 2D scale metric}
\end{figure}

Furthermore, we deployed the trained controller on \fakest{ a larger swarm with 100 agents.} larger \revision{swarms to test its scalability. Each of these swarms are uniformly initialized in a ball around the origin, with the same robot density as the training data. Fig. \ref{fig: 2D scale metric} shows the controller performance on swarms of sizes from 10 to 90. It can be observed that $amd$ remains largely consistent, demonstrating that collision avoidance is effective and cohesion is in place even as the swarm size increases. Although $avd$ degrades as the swarm size increases, it remains low enough that some velocity alignment is achieved. As a qualitative illustration,} Fig. \ref{fig: 2D scaling} shows \revision{six snapshots of a swarm of 100 robots using the learnt controller. Although qualitatively velocity alignment can be observed in the predicted trajectories from the snapshots, the global behavior is different from the simulation. This is because the simulation uses the \textit{global} controller while our prediction uses the decentralized controller learnt from the 10-agent data. In other words, the predictions are the best effort to mimic the centralized 100-agent swarm using the learnt decentralized controller.}\fakest{that the learnt controller generalizes to swarms with more agents and flocking emerges after about 800 steps.} We do note that with some initialization, the \revision{predicted} 100-agent swarm tends to split into subswarms. This is not unexpected since stability of the original controller is only guaranteed under certain conditions \cite{tanner2003stable,tannerpartii}.

\begin{figure}
    \centering
    \vspace*{0.1cm}{\includegraphics[width=0.44\textwidth]{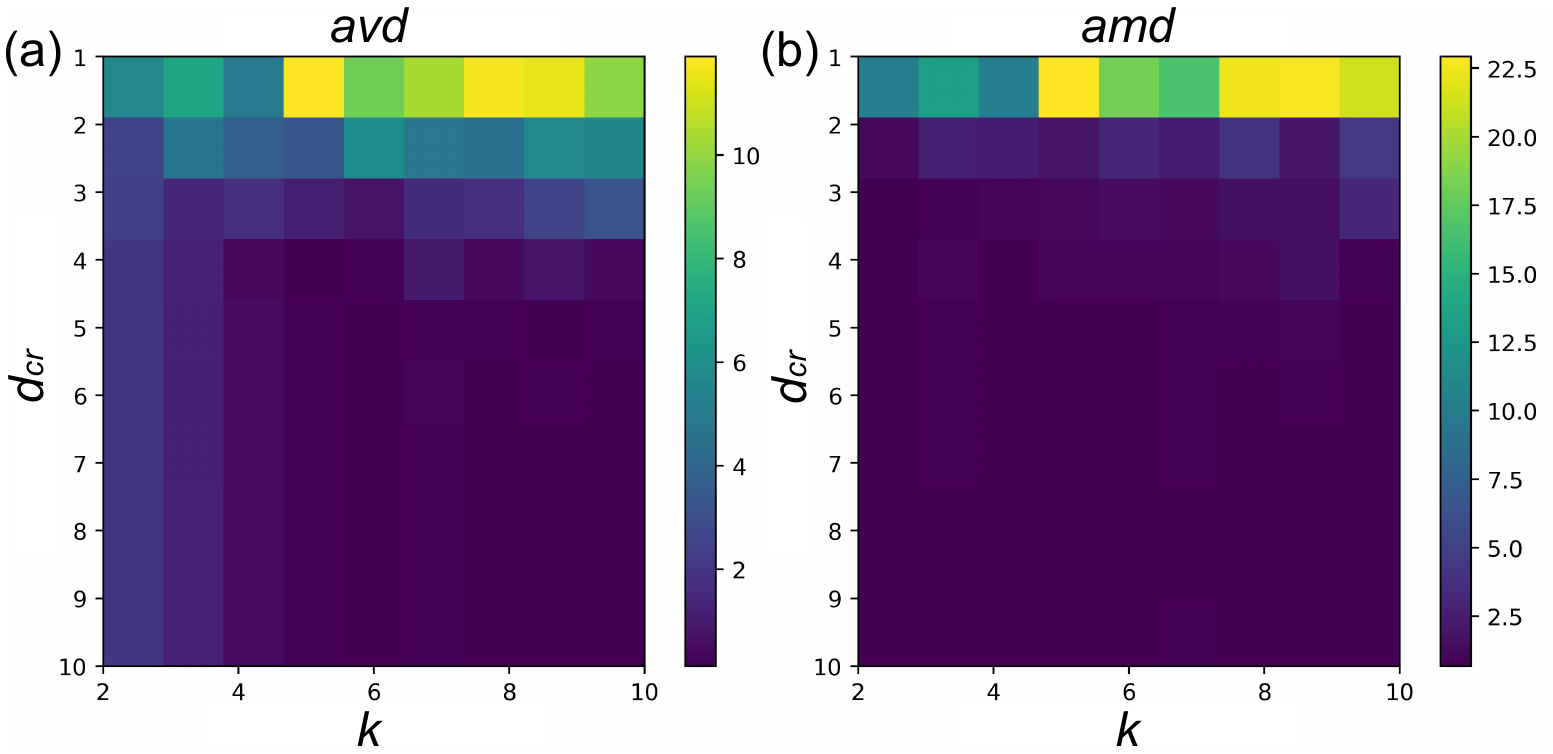}}
    \caption{\revision{Grid search on (a) the average velocity difference ($avd$) and (b) average minimum distance ($amd$) to a neighbor using different communication radii and number of active neighbors. The grid values are computed as the average over trajectories using 20 different initial conditions.}}
    \label{fig: 2D grid search}
\end{figure}

\revision{We further conducted analysis on the hyperparameters $d_{cr}$ and $k$ with respect to 2D flocking. Grid searches are performed on both $avd$ and $amd$ by varying $d_{cr}$ and $k$. For each grid, the average of the last 10 steps of a 2000-step trajectory are computed for 20 different initial conditions. The average over these 20 different initial conditions is then reported in the grid. It can be observed from Fig. \ref{fig: 2D grid search} that the $avd$ is poor for both small values of $d_{cr}$ and $k$, while $amd$ is largely affected by $d_{cr}$ only. This grid search result agrees with intuition and can help with hyperparameter selection.}

\section{Learning to Flock in 3D}
\label{sec:flocking}
Next, we apply our learning method on the 3D simulation of boids. Boids \fakest{were} \revision{was} introduced to emulate flocking behaviors and led to the creation of \textit{artificial life} in the field of computer graphics~\cite{reynolds1987flocks}. The flocking behavior of boids is more challenging to learn because \revision{(1) they have higher dimensionality, and (2)} their steady state flocking behavior is more complex than the 2D flocking in the previous section when the swarm is confined within limited volume.

\subsection{Simulation in 3D and training}
Boids are simulated based on three rules:
\begin{itemize}
    \item \textbf{cohesion} each boid moves towards the average position of its neighboring boids.
    \item \textbf{alignment} each boid steer towards the average heading of its neighboring boids.
    \item \textbf{separation} each boid steer towards direction with no obstacles to avoid colliding into its neighboring boids.
\end{itemize}
While cohesion and collision avoidance are grouped into one term in the 2D flocking case, boids use two separate terms. Furthermore, the boids in simulation are confined in a cubic space and are tasked to avoid the boundaries.

Boids are simulated in Unity~\cite{SebLague2019}. We follow the default settings with a minimum boids speed of $2.0$, a maximum speed of $5.0$, a communication \fakest{range} \revision{radius} of $2.5$ \revision{(ball)}, a collision avoidance range of $1.0$, a maximum steering force of $3.0$, and the weights of cohesion, alignment, and separation steering force are all set to $1.0$. For obstacle avoidance we set the scout sphere radius as $0.27$, the maximum search distance as $5.0$, and the weight of obstacle avoidance steering force as $10.0$. Boids are simulated in a cubic space with an equal side length of $10$, with each axis ranging from $-5$ to $5$. The boids' positions are randomly initialized within a sphere of radius $5$ centered at origin, and their velocities vectors are randomly initialized with a constant magnitude. 

Unity can log both the positions and velocities of boids. However, to make the learning task more challenging, we \fakest{normalize the velocities before including them in the training data, \textit{i.e.}, only the positions and orientation of the boids are given} \revision{only use the positions and orientations of the boids for training}. For a swarm of 10 boids\revision{,} we simulate \fakest{6} \revision{22} trajectories, each with a total of 1700 steps. We discard the first 10 time steps to remove simulation artifacts (There are 'jumps' in the first few steps of simulation) and only use the remaining 1690 steps. We use 2 trajectories for training and the remaining \fakest{4} \revision{20} as the testing data. Zero-mean Gaussian noise with variance 0.01 is added to the training trajectories.

The training model follows \eqref{eqn: final controller}.  The controller $\hat{u}_{\bftheta}$ uses a one layer neural network with 128 hidden units and a hyperbolic tangent activation function. In addition to collision avoidance, we also \revision{include the knowledge for avoiding} \fakest{avoid} the boundaries of the cubic space. This is implemented by \fakest{avoiding} \revision{treating} the closest \fakest{three} point\fakest{s on the boundaries} \revision{on each boundary as an obstacle} \fakest{in the three axes} at any given time. Collision and obstacle avoidance use different \fakest{trained} gains, both of which are defined as $\lambda = \phi^2$, \revision{where $\phi$ is trained}. We further assume an information delay of $1$.

\subsection{Evaluating flocking in 3D}
\textbf{Average minimum distance to a neighbor} \revision{($amd$)} from \eqref{eqn: amd} is also used for 3D flocking to measure the cohesion between robots. However, $avd$ is not a good metric for evaluating flocking in 3D for two reasons: (1) boids only \fakest{align their velocity} \revision{achieve velocity alignment} with the local flockmates because of the presence of obstacles, and (2) boids form subswarms. As a result, \revision{global} velocity alignment is often not achievable at steady state flocking. We instead compare the \textbf{Proper orthogonal decomposition (POD) modes} of the true and predicted trajectory to check how similar the energy distributions are in their respective dynamics. \revision{Built on singular value decomposition,} POD is a model order reduction technique \revision{for nonlinear high-dimensional dynamical systems}. \fakest{which} \revision{It} first decomposes the trajectory of a system into \revision{orthonormal} modes\fakest{based on their \textit{energy} \cite{holmes_lumley_berkooz_1996}.}\revision{, and then truncates the system by selecting from these modes to form a low-rank basis that captures the most \textit{energy} of the system \cite{holmes_lumley_berkooz_1996}.} Systems with similar dynamics should have similar distributions of POD modes \fakest{modes} \revision{when their energies are arranged in descending order. To measure the shift in the distribution of POD modes between the predicted trajectories and ground truth, we further employ the Kullback-Leibler divergence (KLD), which measures the statistical distance between probability distributions \cite{Joyce2011}. Together, we first perform POD on trajectories to find the distribution of their energies. Then we apply KLD on the normalized POD distribution to quantitatively measure the shift in this distribution from the ground truth. We name this metric POD-KLD.} \revision{To generate trajectory predictions, we directly use the learnt controller for velocity control of the swarm.}

\begin{figure}
    \centering
    \vspace*{0.2cm}\includegraphics[width=0.45\textwidth]{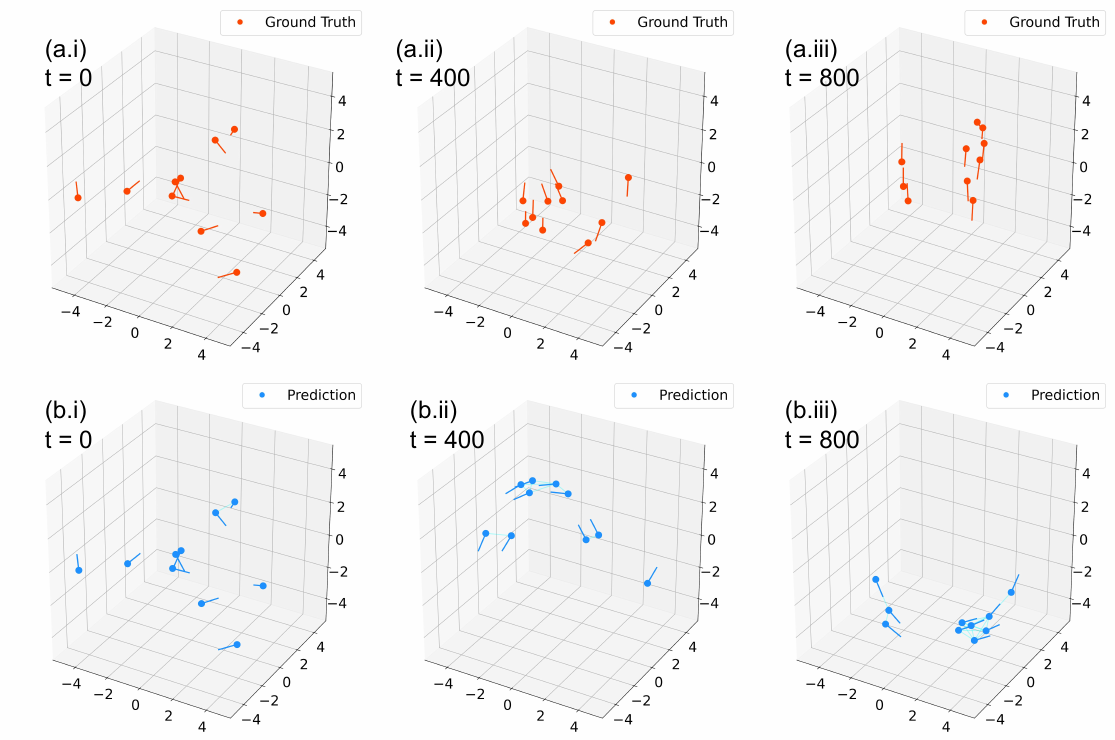}
    \caption{\revision{Predicted trajectories of 10 robots using the learnt controller ($d_{cr} = 2, k = 6$) and the same initialization as the testing trajectory (ground truth).} The subfigures (a.i)(a.ii)(a.iii) show snapshots of the ground truth trajectory at $t=0, 400, 800$, and (b.i)(b.ii)(b.iii) show the eventual flocking and the formation process of subswarms at $t=0, 400, 800$. The light blue lines connect the neighbors in the swarm.}
    \label{fig: 3D pred}
\end{figure}

\begin{figure}
    \centering
    \vspace*{0.2cm}{\includegraphics[width=0.48\textwidth]{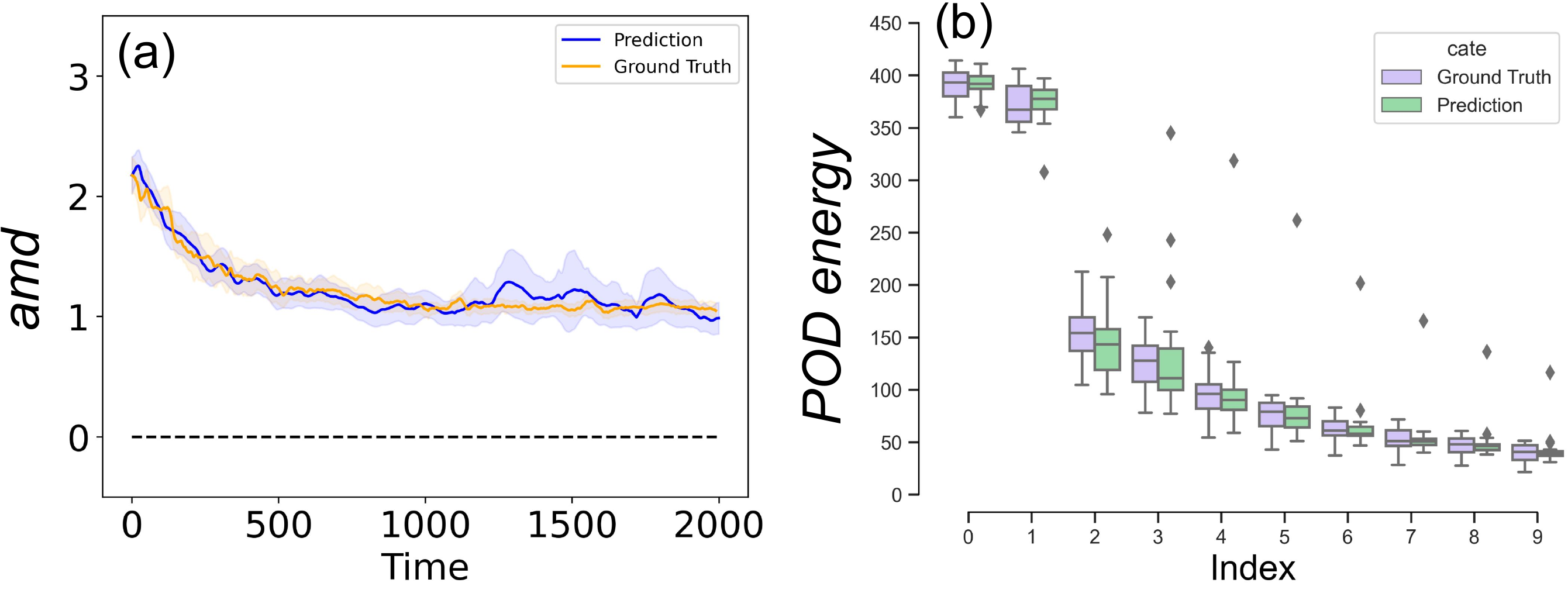}}
    \caption{The metrics for the learnt 3D controller ($d_{cr} = 2, k = 6$). (a) \revision{Average minimum distance to a neighbor of} the predicted trajectory converges, and (b) the distribution of the first 10 POD modes of the predictions and ground truth are similar. The $95\%$ confidence intervals \revision{of $amd$} are based on 20 sets of testing trajectories.}
    \label{fig: 3D metric}
\end{figure}

\begin{figure}
    \centering
    \vspace*{0.1cm}{\includegraphics[width=0.46\textwidth]{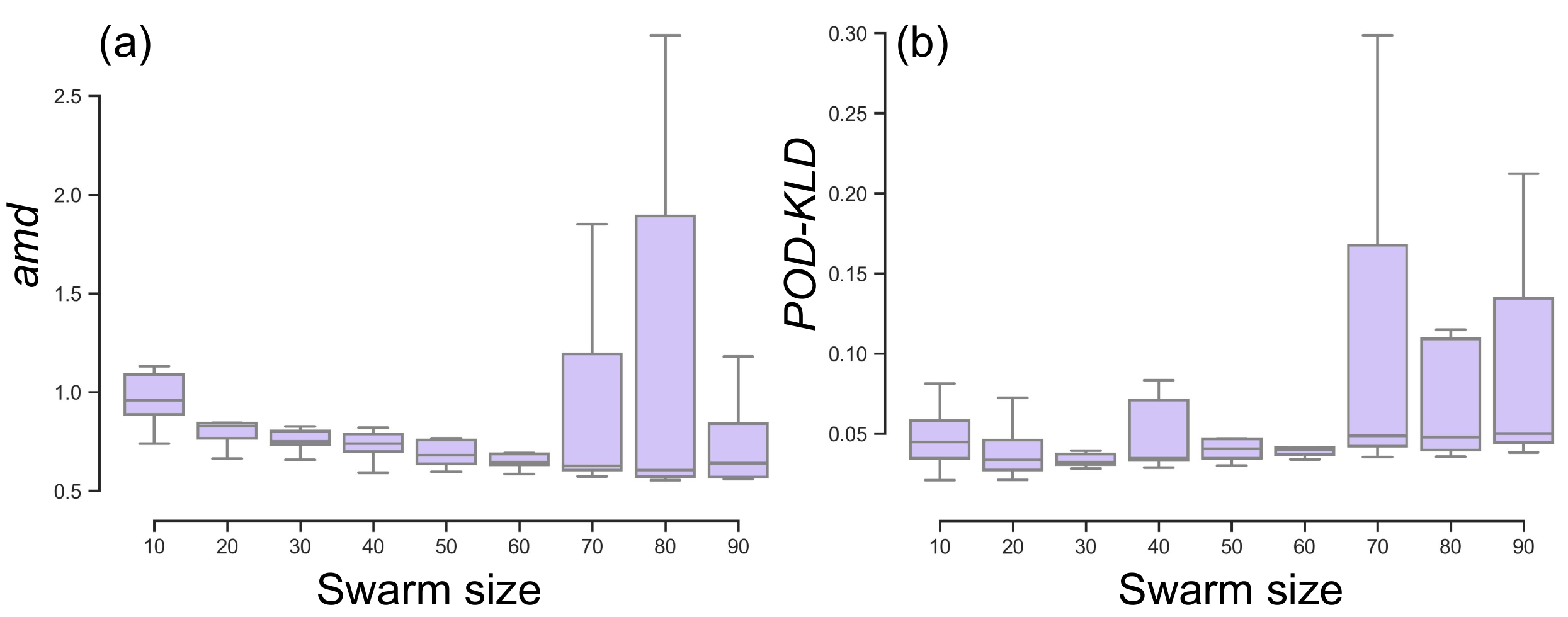}}
    \caption{\revision{Box plot of (a) average minimum distance to a neighbor ($amd$), and (b) POD-KLD of trajectories generated by a learnt controller on swarms of different sizes in 3D. For each swarm size, the box represents the statistics of 15 runs using different initial conditions.}}
    \label{fig: 3D scale metric}
\end{figure}

\begin{figure}
    \centering
    {\includegraphics[width=0.45\textwidth]{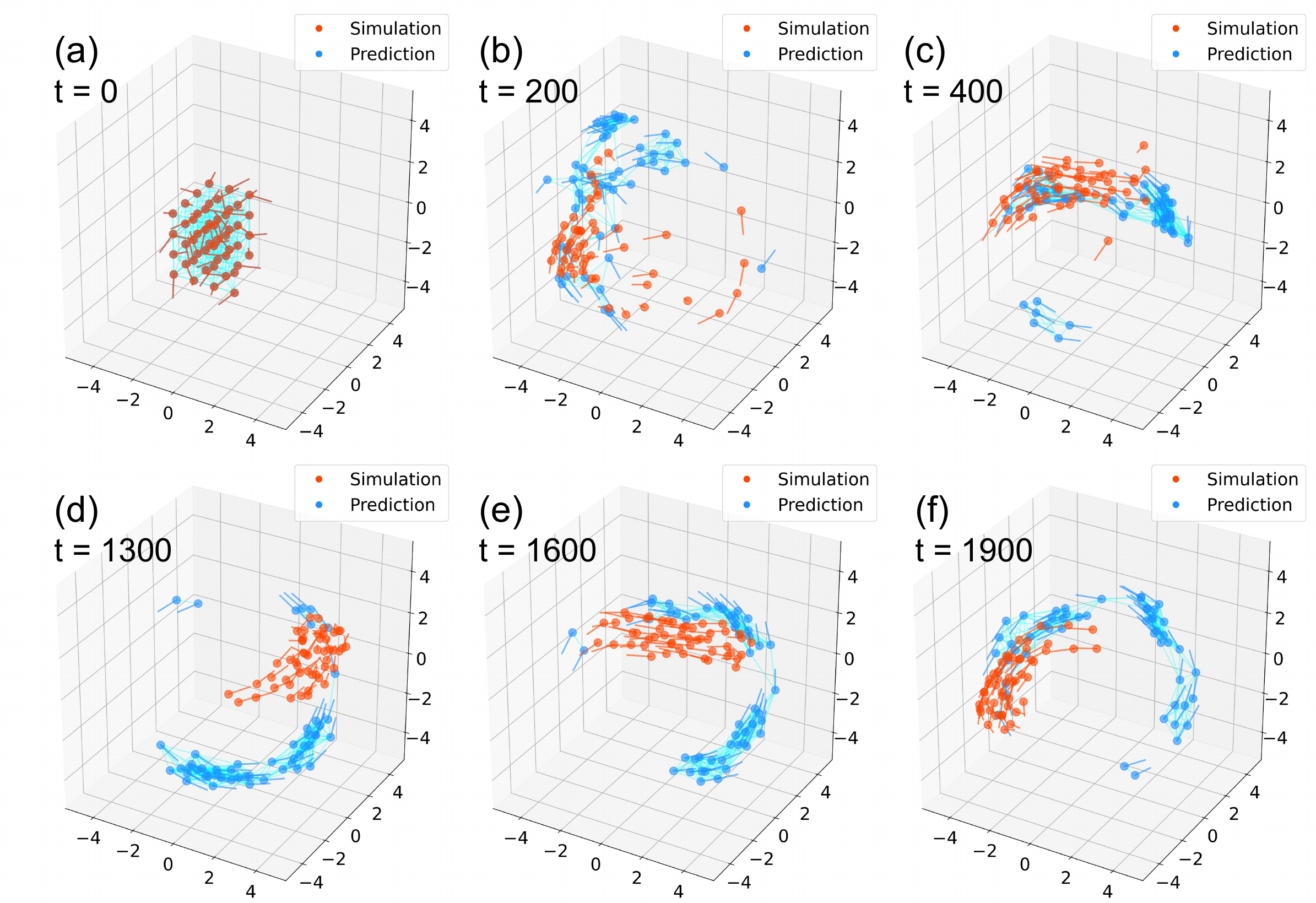}}
    \caption{The flocking of 50 robots using the learnt controller ($d_{cr} = 2, k = 6$) with uniformly initialized positions. The subfigures (a)(b)(c)(d)(e)(f) show the snapshots of the swarm at $t=0, 200, 400, 1300, 1600, 1900$ respectively. The light blue lines connect the neighbors in the swarm.}
    \label{fig: 3D scaling}
\end{figure}

\begin{figure}
    \centering
    {\includegraphics[width=0.5\textwidth]{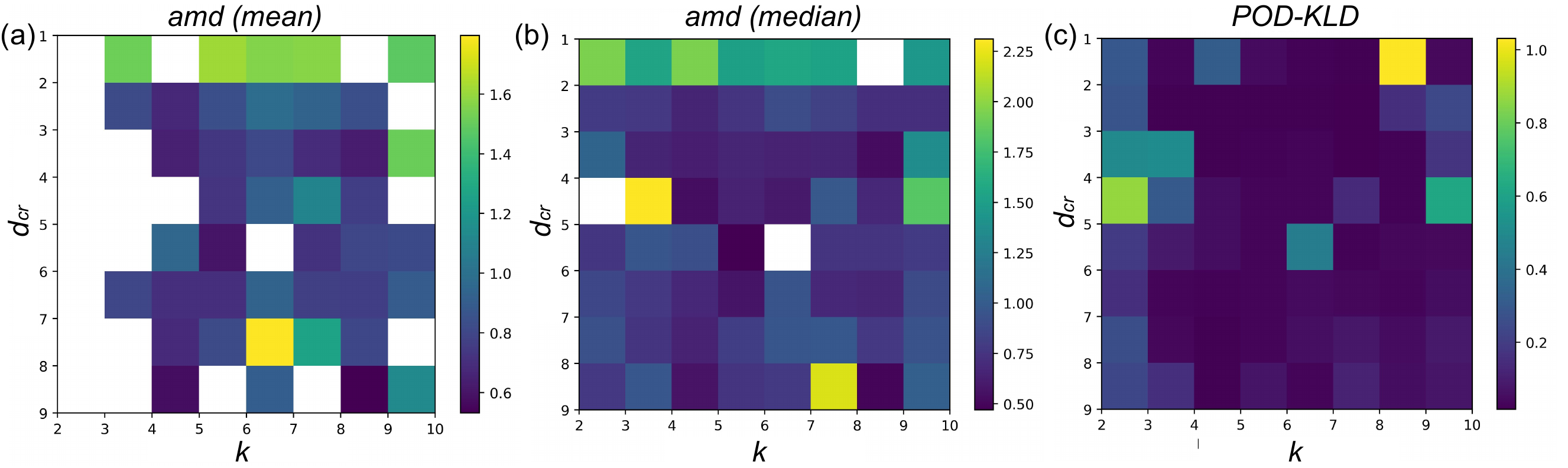}}
    \caption{\revision{Grid search on (a) the mean of average minimum distance to a neighbor ($amd$), (b) the median of $amd$, and (c) the mean of KL divergence of the POD modes using different communication radii and number of active neighbors. The grid values are computed over trajectories using 20 different initial conditions. For $amd$, the grids in white represent values greater than 3. It can be observed that $k$, the number of neighbors to keep has large influence on the metrics.}}
    \label{fig: 3d grid search}
\end{figure}

\subsection{3D Results}
Fig. \ref{fig: 3D pred} \fakest{shows the testing data and the swarm trajectories generated using the learnt controller. It uses a communication radius of 2 and the number of active neighbors is 6. The robots are initialized using the same initial states as the testing trajectory.} \revision{shows a qualitative comparison between the testing data and the trajectory generated by a controller trained with $d_{cr} = 2$ and $k=6$ using the same initial conditions.} The predicted \fakest{trajectories} \revision{trajectory} shows the formation of subswarms during steady state flocking similar to that of the testing trajectory. \fakest{Note that when the robots are initialized closer to each other, they} \revision{Empirically the robots} are more likely to form a single swarm \fakest{during} \revision{at} steady state \fakest{flocking} \revision{when the robots are initialized closer to each other}. The metrics for the learnt controllers are shown in Fig. \ref{fig: 3D metric}. It can be seen that group cohesion is achieved as both the predicted and true swarm show similar trends for $amd$. Furthermore, the distributions of POD modes between the prediction and testing data are similar, indicating similar dynamics. 

\fakest{Lastly, we apply the learnt controller on a larger swarm with 50 agents. Fig. \ref{fig: 3D scaling} shows the emergence of flocking behavior at about $t=400$.}
\revision{We also test the scaling ability of the learnt controller on larger swarms of sizes ranging from 10 to 90. Each of these swarms are uniformly initialized in a ball around the origin, with the same robot density as the training data. Fig. \ref{fig: 3D scale metric} shows the metrics on trajectories of different swarm sizes using the same learnt controller. It can be observed that the trend for $amd$ is better than the 2D case as swarm size increases. This can be explained by the fact that robots are confined in a cubic space and do not travel too far from each other. Fig. \ref{fig: 3D scaling} shows comparison between predictions and simulation when there are 50 robots. Notice that our prediction forms subswarms with this size. This may also occur in simulations of 50 agents in Unity. Both qualitatively and quantitatively, the controller learnt in 3D scales better than that in 2D. One possible reason is that the 3D simulation itself is decentralized, while the ground truth controller in 2D is centralized. Hence, the predicted trajectories of a larger swarm in 3D is closer to that in simulation.}

\revision{A grid search is also performed on the hyperparameters $d_{cr}$ and $k$ for 3D flocking. The results are shown in Fig. \ref{fig: 3d grid search}. While a small $k$ leads to poorer metrics, communication range $d_{cr}$ does not affect the metrics as significantly as in the 2D case. This may be due to fact that the swarm in 3D are confined in a fixed volume, and therefore the higher density of robots leads to higher chance for the robots to come within each other's communication range even if their communication range is small. Additionally, it can be seen that not all trained models converge. Especially for small $k$, cohesion may not be achieved in the resulting swarm. Visual inspections reveal that these instances correspond to when robots overcome the obstacle avoidance potential function and leave the cubic space. Since such singular cases increase the average $amd$ dramatically, to better assess the performance we also plot the medians of $amd$ in Fig. \ref{fig: 3d grid search}. Another observation is that the performance degrades slightly for large $k$. This can be explained by the increase in the number of neural network parameters -- an increment of 1 in $k$ correspond to an increase of $256$ parameters as the input size increases. Since the training data and training time are unchanged, a larger neural network may tend to underfit.}

\section{Discussion}
Our experiments show that the model proposed in \eqref{eqn: final controller} is able to learn flocking in both 2D and 3D \revision{using appropriate hyperparameters $d_{cr}$ and $k$}. The choices of $d_{cr}$ and $k$ and the corresponding learnt controllers can inform how the extent of decentralization can affect flocking behavior in robot swarms. \revision{Furthermore, we note that the collision avoidance strategy which we used as knowledge does not guarantee collision-free trajectories. This is evident in Fig. \ref{fig: 3d grid search} where robots using some trained controllers leave the confined box. However, the use of this collision avoidance strategy demonstrates the flexibility of our proposed framework for embedding known knowledge about single-robot dynamics, and users are free to incorporate any knowledge including but not limited to collision avoidance strategies.} \fakest{We perform a grid search using different $d_{cr}$ and $k$ for 2D flocking. To evaluate the performance of flocking in 2D in the grid search, we use a single number metric, namely the velocity variance which is given by $C = \frac{1}{n}\sum^m_{i=1}\sum^n_{j=1}\left\|\mathbf{v}_{j}(t_i)-\frac{1}{n}\left[\sum^n_{k=1}\mathbf{v}_{k}(t_i)\right]\right\|^2.$ This metric measures the spread of robot velocities throughout time. It should be small if robots reach concensus in their velocities. Fig. \ref{fig: 2D grid search} shows the grid search result. We observe that for small proximity radii or small number of active neighbors, the velocity variance tends to be large. This matches our intuition as a small $d_{cr}$ or $k$ means that each robot can communicate with fewer neighbors and therefore has less information to act upon in order to stay connected and flock as a swarm.}

\section{Conclusion and Future Work}
We have introduced an effective machine learning algorithm for learning to swarm. Specifically, we applied the algorithm to flocking swarms in 2D and 3D respectively. In both cases, the learnt controllers are able to reproduce \revision{global} flocking behavior similar to the ground truth. Furthermore, the learnt controllers can scale to larger swarms to produce flocking behaviors. We have shown the effectiveness of knowledge embedding in learning decentralized controllers, and demonstrated the feasibility of learning swarm behaviors from state observations alone, distinguishing our work from prior works on imitation learning. For future work, we plan to \revision{learn from real-world data, and} implement the learnt controllers on physical robot platforms\fakest{ to emulate swarming behaviors}. In addition, we hope to employ neural networks with special properties to derive stability guarantees for the learnt controllers.






\bibliographystyle{IEEEtran} 
\bibliography{IEEEabrv, refs}

\addtolength{\textheight}{-12cm}

\end{document}